\documentclass{article}

\usepackage{arxiv}

\usepackage[utf8]{inputenc} % allow utf-8 input
\usepackage[T1]{fontenc}    % use 8-bit T1 fonts
\usepackage{hyperref}       % hyperlinks
\usepackage{url}            % simple URL typesetting
\usepackage{booktabs}       % professional-quality tables
\usepackage{amsfonts}       % blackboard math symbols
\usepackage{nicefrac}       % compact symbols for 1/2, etc.
\usepackage{microtype}      % microtypography
\usepackage{lipsum}		% Can be removed after putting your text content
\usepackage{graphicx}
\usepackage{doi}
\usepackage{float}
\restylefloat{table}
\usepackage[square,sort,comma,numbers]{natbib}

\title{ViT-Inception-GAN for Image Colourising}

\author{ \href{https://orcid.org/0000-0001-5519-3071}{\includegraphics[scale=0.06]{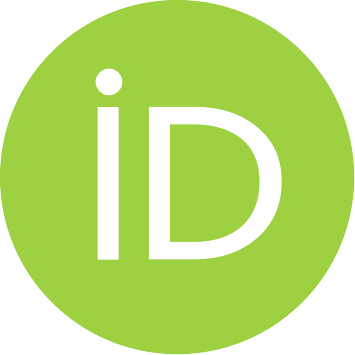}\hspace{1mm}Tejas Bana}\\
	Department of Information Technology\\
	D.Y Patil College of Engineering\\
	Akurdi, India \\
	\texttt{tejasbana@gmail.com} \\
	\And
	\href{https://orcid.org/0000-0002-1553-7754}{\includegraphics[scale=0.06]{orcid.pdf}\hspace{1mm}Jatan Loya} \\
	Department of Computer Engineering\\
	Vishwakarma Institute of Technology\\
	Pune, India \\
	\texttt{jatan.loya18@vit.edu} \\
	
	\And
	\href{https://orcid.org/0000-0001-9100-4565}{\includegraphics[scale=0.06]{orcid.pdf}\hspace{1mm}Siddhant Kulkarni} \\
	Department of Computer Science\\
	BITS Pilani\\
	Hyderabad, India\\
	\texttt{f20180185@hyderabad.bits-pilani.ac.in} \\
	
}

\renewcommand{\shorttitle}{ViT - Inception - GAN for Image Colourising}

\hypersetup{
pdftitle={ViT - Inception - GAN for Image Colourising},
pdfsubject={},
pdfauthor={Tejas Bana, Jatan Loya, Siddhant Kulkarni},
pdfkeywords={Image Colourising, GAN, Vision Transformer, Deep Learning, Inception-v3},
}

\begin{document}
\maketitle

\begin{abstract}
	Studies involving colourising images has been garnering researchers' keen attention over time, assisted by significant advances in various Machine Learning techniques and compute power availability. Traditionally, colourising images have been an intricate task that gave a substantial degree of freedom during the assignment of chromatic information. In our proposed method, we attempt to colourise images using Vision Transformer - Inception - Generative Adversarial Network (ViT-I-GAN), which has an  Inception-v3 fusion embedding in the generator. For a stable and robust network, we have used Vision Transformer (ViT) as the discriminator. We trained the model on the Unsplash and the COCO dataset for demonstrating the improvement made by the Inception-v3 embedding. We have compared the results between ViT-GANs with and without Inception-v3 embedding. 
\end{abstract}

% keywords can be removed
\keywords{Image Colourising \and GAN \and Vision Transformer \and Deep Learning \and Inception-v3}

\section{Introduction}

Colourisation is a technique that adds colour to grayscale images. Grayscale images often referred to as "black and white" images commonly present in older multimedia. Chromaticity is the quality of colour irrespective of its luminance. Hence accurately adding chromatic data to images has become a popular focus for research.

With the recent advances in machine learning, researchers have developed a sophisticated network known as Convolutional Neural Networks (CNNs) \cite{lecu:yosh:geof}, which have surpassed traditional machine learning techniques that involved feature engineering. Today, CNN's are used in various tasks like image recognition, autonomous driving, video analysis, drug discovery and many more.

Generative Adversarial Network (GAN) \cite{good:jean:mehdi} by Goodfellow et.al is a framework where the generator is put in competition against the discriminator to find if a sample is from model distribution or data distribution. This enables the generation of new data with the same statistics and distribution as the training data. Applications of GANs can range from generating art to reconstructing 3D models of objects from images \cite {wu:ji:ch}. GAN, more precisely, StyleGAN \cite{kar:ter:sam}, which Nvidia developed, has been used to create indistinguishable fake human faces. Some researchers have also used StyleGAN for other applications like colouring images.

Since then, many colourisation methods have been proposed, including but not limited to new model architectures [\cite{chromagan}, \cite{llang}, \cite{text2color}, \cite{pixsem}, \cite{sar}, \cite{deepex}]. Though making seemingly promising progress in colourising, these methods still have some drawbacks. These drawbacks include requiring extensive computational resources and a large dataset for training. Hence, we aim to improve the model performance on a smaller dataset while requiring less computational resources. Our proposed method is fully end-to-end and does not require any user inputs or hints.

Our main contribution in this research work are:
	\begin{enumerate}
		\item Proposed two novel GAN architectures Vision Transformer - Inception - Generative Adversarial Network (ViT-I-GAN) and Vision Transformer - Generative Adversarial Network (ViT-GAN)
		\item Proposed Vision Transformers as a discriminator for the training of GANs
		\item Demonstrating improvement in GAN performance where dataset is limited by fusing Inception-v3 \cite{sze:chr:wei} embedding into the generator.
		\item Identifying and integrating appropriate architectural features and developing training procedure on limited dataset which leads to a boost in GAN performance for image colourisation. 
	\end{enumerate}

\section{Background}
Researchers have undertaken extensive work which involves colourising images to colourise images. Earlier, this involved laborious work since images had to be colourised manually using tools like Adobe Photoshop, which would take up to months to colourise a single image. Colourising a face could take up to 20 layers of green, blue and pink to bring out a satisfactory result. This time has been cut down dramatically with modern machine learning models, which can outperform manual colourising.

\subsection{GANs for Image-to-Image Translation}

The idea of Deep Convolutional Generative Adversarial Networks (DCGANs) \cite{rad:ale:luk}, StyleGAN \cite{kar:ter:sam}, Pix2Pix GAN \cite{pixsem} is to have a generator that is trained for an image to image translation to generate desired images from input data and a discriminator which is trained to discriminate between original and generated images. The generator and discriminator compete with each other, eventually improving the generator's capability for generating authentic looking images.

We can change the input to GAN and use it for various tasks like image colourising. Training GANs is a volatile task that is sensitive to model architecture, specific implementation and hyper-parameter tuning.

Traditionally in GANs, noise is sent as the input to the generator of the network, and then it generates required data from the given noise. Whereas, for image to image translation, the noise is replaced by the features extracted from an image through a CNN encoder which is sent as the input to the decoder, which consequently gives the desired output. 	

\subsection{Vision Transformer}

Vision Transformer (ViT) \cite{dos:ale:luc} was proposed by Dosovitskiy et al., self-attention based architectures, more importantly, Transformers were the top choice for natural language processing (NLP). Transformers have enabled a high degree of efficiency, scalability and speed, which allows us to train models over a huge number of parameters.

Conventionally, CNNs have been the go-to for most computer vision tasks. Inspired by the immense success of Transformers in NLP, researchers have implemented transformers for image classification. For doing this, images are split into patches, and a sequence of linear embeddings is sent as an input to the Transformer. They’ve also concluded that large scale training outperforms inductive bias, which CNNs are known to have. ViT is on-par or beats state-of-the-art CNNs while using fewer parameters and compute resources.

\section{Related Work}
ChromaGAN \cite{chromagan} is a fully automatic end-to-end adversarial approach combined with semantic and perceptual information to colourise images. The generator processes a grayscale image of 224 x 224 as the input. The generator is made up of two branches, one branch returns the chrominance information, and the other branch returns the class distribution vector as a result. The authors trained the model on 1.3M images from a subset of images taken from ImageNet \cite{imgnet}. Since the dataset is extensive, a single epoch took around 23 hours on a NVIDIA Quadro P6000 GPU. For evaluating results, the authors have used peak signal to noise ratio (PSNR) and Naturalness for quantitative evaluation. 

In the paper Learning to Color from Language \cite{llang}, the authors have presented two language-conditioned colourization architectures, which they claim performs better than language-agnostic versions. The authors experimented with CONCAT \cite{concat} and FILM \cite{film} networks but decided to settle on the FILM network given its small number of parameters. Their CNN and the two language-conditioned architectures were trained on  82,783 images of the MS-COCO \cite{lin:tsu:mic} test dataset and where each image has five different crowdsourced captions. Since training the model on MSCOCO resulted in poor colourization, they decided to initialize all convolutional layers with a CNN pre-trained on ImageNet\cite{imgnet}. They ran three human evaluations on the Crowdflower platform to evaluate overall quality, plausibility and how well they condition their output on language. The model accuracy was evaluated at a downsampled resolution of 56 x 56, and the predictions were upsampled to 224 x 224 for human experiments.

Text2Colors \cite{text2color} model consists of two conditional GAN: the text-to-palette generation networks (TPN) and the palette-based colourization network (PCN). The TPN provides the relevant colour palette based on the semantics of the input text. PCN colourizes a grayscale image using the colour palette generated by the TPN. The authors have also created a Palette-and-Text (PAT) dataset containing 10,183 text and 5 colour palette pairs. This allowed them to train their models for predicting colour palettes semantically consistent with their text inputs. The TPN was trained on their PAT dataset and the PCN was trained on CUB-200-2011 (CUB) \cite{cub} and Imagenet ILSVRC Object Detection (ImageNet) \cite{imagenet} dataset. Palette evaluation and user study were done as a part of the quantitative evaluation. In the former, they demonstrated diversity and multimodality of the model, and in the latter, 53 participants gave their inferences, where they claim the users preferred the palettes generated by their model over palettes created by a human.

In Pixelated Semantic Colorization \cite{pixsem} , the authors have proposed to exploit pixelated object semantics to guide image colourization. An autoregressive model is adopted to use pixelated semantics for colourization. They have used pixelated semantic embedding and a pixelated semantic generator to integrate object semantic in the colourization model. The output is obtained by fusing multi-scale features. The network is trained on Pascal VOC2012 \cite{pascal} and COCO-stuff  \cite{lin:tsu:mic} dataset, where 10,582 images were used for training and 1449 images in the validation set for testing. Grayscale input images are rescaled to 128 x 128 to reduce computation. Mean intersection over union (Mean-IoU), peak signal to noise ratio (PSNR), root mean squared error (RMSE), and Naturalness were used for quantitative comparison with state-of-the-art models.

To colourize Synthetic Aperture Radar (SAR) images, Wang et al. \cite{sar} proposed SAR-GAN. Synthetic aperture radar (SAR) is a coherent radar imaging technology that is capable of producing high-resolution images of targets and landscapes. 
SAR-GAN is based on a cascaded network of convolutional neural nets (CNNs) for despeckling and image colourization. It consists of three components: despeckling sub-network, colourization sub-network and generative adversarial learning. Despeckling sub-network is used to restore a clean image from a noisy observation. The colourization sub-network then transforms the despeckled image into a visible image.

Deep exemplar-based colourization \cite{deepex} transfers the colours from a reference image to the grayscale one. The aim is to provide diverse colours to the same image and does not focus on colourizing images naturally. The model consists of two subnetworks: Similarity subnetwork and Colorization subnetwork.
 The similarity subnetwork takes the target and reference luminance channels aligned before via Deep Image Analogy \cite{analogy}. The authors have used the standard features of VGG19 \cite{vgg19}. The similarity subnet computes a bidirectional similarity map using discrete cosine distance. The colourization subnetwork concatenates the calculated similarity maps, chrominance channels, and the grayscale image. The structure of the colourization subnet was inspired by U-Net \cite{unet}.
dataset used for training was based on ImageNet dataset involving 700 classes out of the total 1,000 classes.by sampling approximately 700,000 image pairs from 7 popular categories: food (5\%), people (20\%), artifacts (5 \%), scenery (25\%), ,animals (15\%), transportation (15\%) and plants (15\%). The loss function used was l2, which is a combination of chrominance channels and perceptual loss. Adam optimizer was used with a learning rate of 10-3 for ten epochs, reduces by 0.1 after 33\% of training in the Caffe \cite{cafe} framework was used to train the network.

\section{Proposed Method}

Conventionally, images are represented in Red-Green-Blue (RGB), where there is one layer for each colour. Images are represented in a grid of pixels, where the pixel value ranges from 0 (black) - 255 (white). We have considered the CIE L*a*b* colour space, where L stands for Lightness, and a \& b for the colour spectra green-red and blue-yellow. The RGB input image of size H x W firstly converted to L*a*b* colour space, and the L (Luminance) layer is the input for the model; this gives the semantic information of the image to the model, including but not limited to the objects and their textures. The task is to predict the other two colour channels. By coalescing luminance and predicted colour channels, the model warrants a high level of detail in the final colourised images. For colourising images, the neural network creates a correlation between grayscale input images and coloured output images. Our goal is to determine this link as accurately as possible and observe the effect of additional features produced by InceptionNet-v3.

\subsection{Preprocessing}

In the CIE L*a*b* colour space, the value of L* generally ranges from 0 to 100, and a* \& b* range from -128 to 128.  As a result of using vision transformers in our model, the input image's size to the classifier must be fixed. We have resized input images to 256 x 256 for the encoder and discriminator. The input size for our network is not restricted to the input size of the pre-trained Inception-v3. Therefore, we have triplicated the luminance channel L to create a three channel grayscale image and have resized it to 299 x 299 for Inception-v3. We have normalised the pixel values for the generator between [-1, 1] and input values for the Inception-v3 model within the interval of [0, 1]. 

\subsection{Architecture}

Our models' architecture is inspired by U-Net architecture \cite{unet} and the usage of fusion layer which was proposed by S. Iizuka et al\cite{iiz:sat:edg} in the autoencoder network. Given the luminance information of an image, the model gives its best estimation of alpha and beta components and merges them to give the final colourised output image. We have fetched an embedding of the grayscale image from the last layer of Inception-v3.
The architecture of the generator consists of Encoder, Feature Extractor, Fusion, Decoder. The Luminance channel of the image is given to the Encoder and Feature Extractor; their outputs are merged in Fusion Layer; this information is passed to the Decoder, which outputs a* and b* colour channels of the CIE L*a*b* colour scheme. Then, a* and b* colour channels are merged with the image's Luminance channel to produce a colourised image.

\begin{figure}[!htbp]
  \centering
  \includegraphics[width=0.6\linewidth]{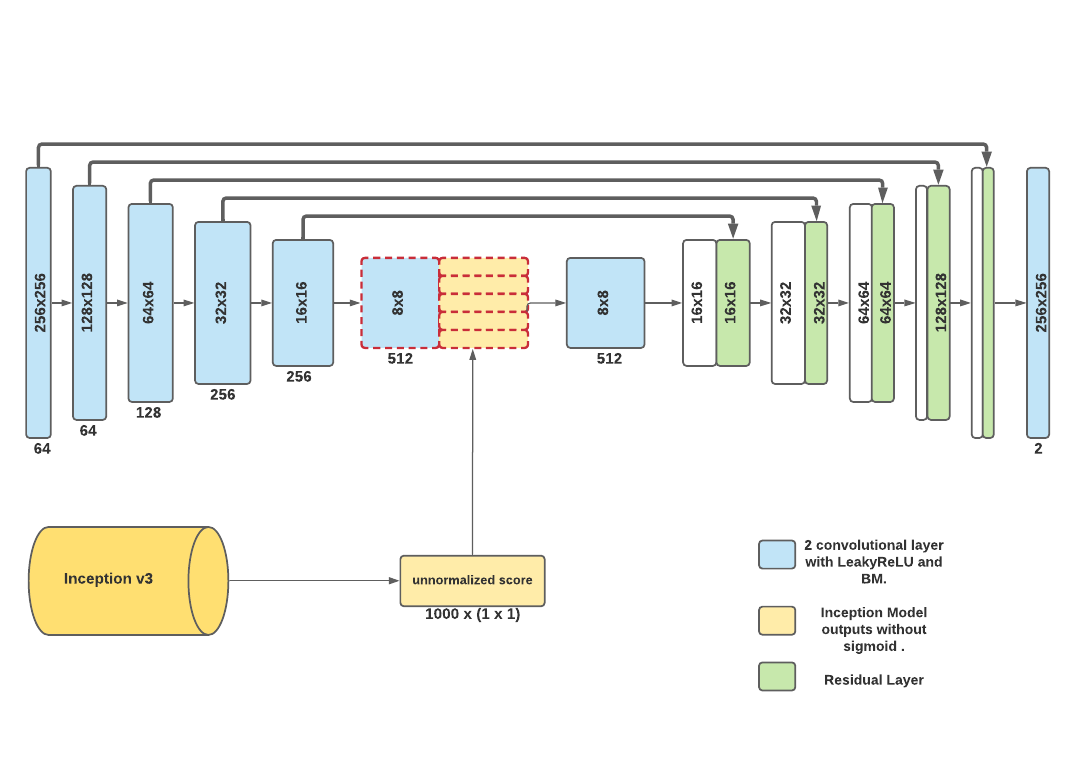}
  \caption{ViT-I-GAN Generator}
  
  \centering
  \includegraphics[width=0.6\linewidth]{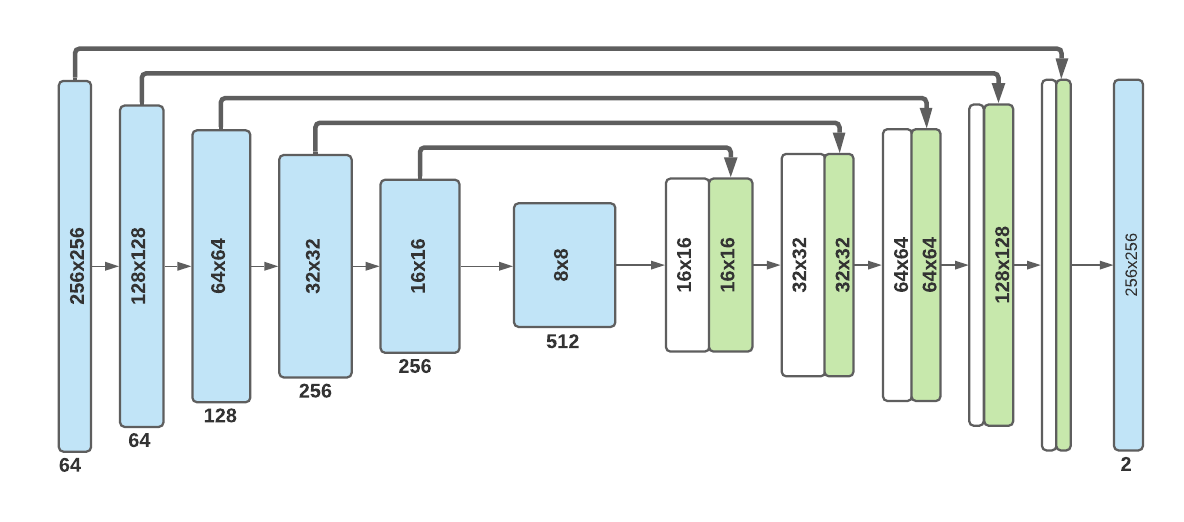}
  \caption{ViT-GAN Generator}
  
  \centering
  \includegraphics[width=0.6\linewidth]{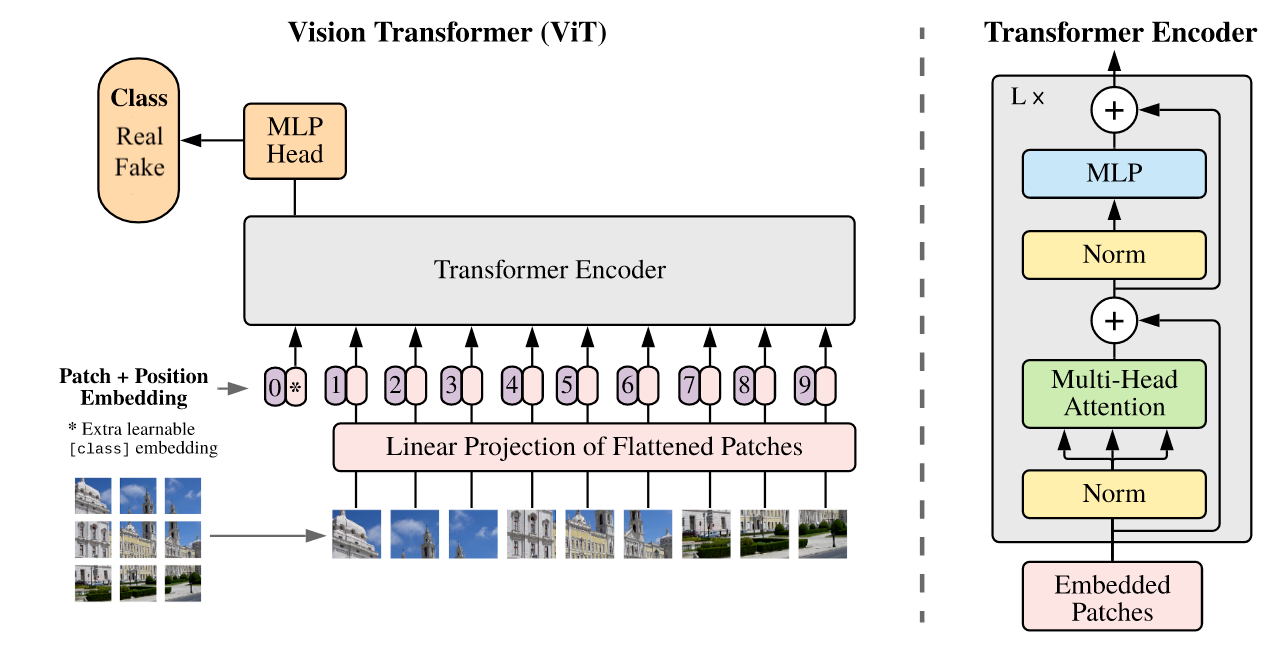}
  \caption{ViT\cite{dos:ale:luc} as Discriminator}
  %\vspace{-2em}
\end{figure}

\subsection{Encoder}

The encoder processes H x W grayscale images and gives H/32 x W/32 x 512 feature representation as the output. The encoder consists of 10 convolutional layers with 5 x 5 kernel, and padding of 2px is applied on each side to maintain the layer's input size.  Furthermore, for downsampling, we have used Average Pooling instead of MaxPooling as it smooths the output image, which halves the dimension of their output and hence reduces required computation. 

\subsection{Feature Extractor}
We have used a pre-trained Inception-v3 model for extracting image embedding. Firstly, the input image is scaled to 299 x 299 px, and normalised input values are received within the interval of [0,1]. Then we pile up these images on themselves to get a three channel image for Inception-v3 dimension criteria. Consequently,  the resulting image is fed to the network and the last layer's output before the softmax function is extracted. This gives us a 1000 ×1×1 embedding.

\subsection{Fusion Layer} 
The fusion layer takes the feature vector from Inception-v3, then replicates it (HxW)/(32 x 32) times and attaches it to the encoder's feature volume along the depth axis. This gives a single volume of the encoded image and the mid-level features of shape H/32 × W/32 × 1512. By replicating the feature vector and concatenating it several times, we ensure that the feature vector's semantic information is uniformly distributed among all spatial regions of the image  \cite{bal:fed:die}. Lastly, we apply 512 convolutional kernels of size 1×1, finally generating a feature volume of dimension H/32 × W/32 × 512.

\subsection{Decoder}
The decoder takes H/32 x W/32 x 512 volume and applies convolution and upsampling layers, which produces a final layer of dimension H x W x 2. Upsampling is performed using Nearest Neighbour Interpolation to ensure the output dimension is twice the input dimension. We use 5 ConvTranspose layers with 3 x 3 kernels followed by LeakyReLU with a negative slope of 0.2.

\subsection{Vision Transformer as a Discriminator}

Vision transformer is one of the most successful applications of Transformer for Computer Vision. A significant challenge of applying Transformers without CNN to images is using Self-Attention between pixels. If the input image size is 256 x 256, the model needs to calculate self-attention for 65K combinations. Also, it is not likely that a pixel at the corner of an image will have a meaningful relationship with another pixel on the other corner of the image. ViT has overcome this problem by segmenting images into small patches (like 32 x 32).
ViT breaks an input image of 256 x 256 into a sequence of patches. Consequently, each patch is flattened into a single vector in a series of interconnected channels of all pixels in a patch, projecting to desired input dimension.
The atom of a sentence is a word, similarly in the case of ViT, a patch is the atom of an image instead of a pixel to tease out patterns efficiently.

The crux of the Transformer architecture is Self Attention. It enables the model to understand the connection between inputs.
ViT combines information across the entire image, even in the lowest layers in Transformers. As quoted in the paper, “We find that some heads attend to most of the image already in the lowest layers, showing that the ability to integrate information globally is indeed used by the model.”

\begin{table}[!htbp]
\centering
\begin{center}
\begin{tabular}{lc}
\hline
\textbf{ Hyperparameter } & \textbf{ Value }\\
\midrule
      Image Size & 256\\
      Patch Size & 32 \\
      Transformer blocks & 6\\
      Heads in Multi-head Attention layer & 16\\
      Dimension of the MLP (FeedForward) layer & 2048\\
      Dropout Rate & 0.1\\
      Embedding dropout rate & 0.1\\
\hline
\end{tabular}
\caption{Hyperparameter of Discriminator}
\end{center}
\end{table}

\subsection{Batch Normalisation}
\cite{iof:ser:chr} This methodology normalises the previous layer's output by subtracting the batch mean and dividing by the standard deviation of the batch. This introduces noise in each layer's output as the estimate of mean and standard deviation is noisy, which in turn reduces overfitting while having a regularising effect. Hence, it is used instead of dropout since both add noise and help minimise overfitting. There is substantial work done to prove that using Batch Normalisation is advantageous for stable GAN training and prohibits the model from collapsing due to poor initialisation. It retains content information by reducing internal covariate shift inside a mini-batch while training.

\section{Experiments}

We have compared ViT GAN and ViT-I-GAN on a small dataset due to limited computing resources. For comparison between the models, we have used the Unsplash dataset, and for fine-tuning, we have used Natural-Color Dataset (NCD dataset) \cite{anw:sae:muh}. VIT-I-GAN was also trained on the COCO dataset \cite{lin:tsu:mic} to test its limits and have a generalised model which can colourise a wide variety of images. COCO dataset encapsulates almost every common object since it covers 172 classes, making the model generalise better. We cannot compare our model with state-of-the-art models because their models are trained on 1.4 million images which is far more than what we could train on.

\subsection{Dataset}

We have used 10,500 images from the Unsplash dataset, which is publicly available. Both models are trained on 10,000 images and are validated and compared on 500 images and NCD dataset, which comprises 723 images of fruits and vegetables. For Fr\'{e}chet Inception Distance (FID) \cite{heu:mar:hub} comparison, we have used 12,000 validation images from the COCO dataset.
\subsection{Training}

We have trained both models for 50 epochs with a batch size of 16 and a total of 31k steps used for training. Inception- v3 model is initialised with pre-trained weights for ImageNet. Adam optimiser is used with learning rate 0.0002 and momentum parameters $\beta_1$ = 0.5  and  $\beta_2$ = 0.9. Adam optimiser is used because it converges faster and provides lower loss values for the generator, which is favourable for training GANs.

The model utilizes a hybrid loss function, developed by combining the pixel-level L1 loss with the adversarial loss.
L1 Loss function is the Least Absolute Deviations (LAD).
LAD is used to minimize the error, which creates a criterion that measures the mean absolute error (MAE) between each element in the input $y_t$ and target $y_p$. 
\begin{center}
L1 Loss Function = $\sum_{i=1}^{n}|y_t - y_p|$
\end{center}
The adversarial loss is calculated by Binary Cross Entropy between the real and generated images.
\centerline{BCE Loss = $-(\frac{1}{N})\sum_{i=1}^{N} y_i \cdot log(\hat{y_i}) + (1-y_i) \cdot log(1-\hat{y_i})$}
\section{Results}

Why fusion layers helps: just by adding more parameters to the generator or making the generator model deeper by adding more layers to it does not improve the generator’s performance and ability to generalise on a limited dataset; hence a parallel network is required, which gives some kind of advantage to the generator, using a pre-trained classifier does work for this purpose since it does not need training for feature extraction hence saving the computational resources, and also the extracted features from the classifier boosts the generator’s ability to perform the task.
Empirically it was observed that the addition of the fusion layer to the generator helps when training is done for a long period of time.

We have compared the results of our models using both qualitative and quantitative evaluation methods. 

\subsection{Comparison on Test Images}
 The test images are from Unsplash dataset and consist of various classes of gray scale images. These are more representative of real world examples where a single image is composed of several objects which makes colourising rather intricate than an image containing just a single object. 
 
 \begin{figure}[!htbp]
\begin{center}
\begin{tabular}{ccc}
	{\includegraphics[width = 1.17in]{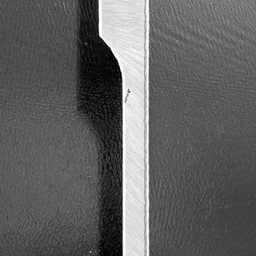}} &
	{\includegraphics[width = 1.17in]{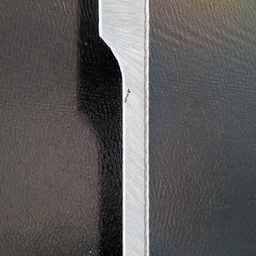}} &
	{\includegraphics[width = 1.17in]{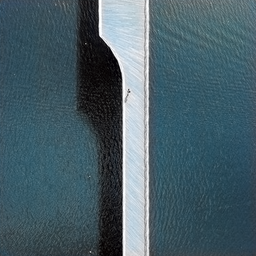}}\\
	{\includegraphics[width = 1.17in]{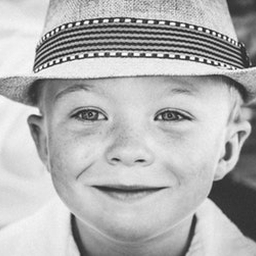}} &
	{\includegraphics[width = 1.17in]{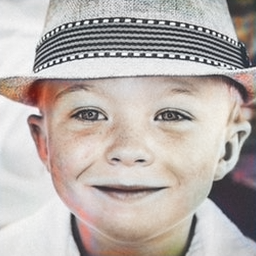}} &
	{\includegraphics[width = 1.17in]{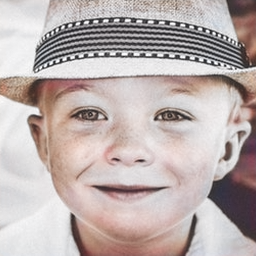}}\\
	{\includegraphics[width = 1.17in]{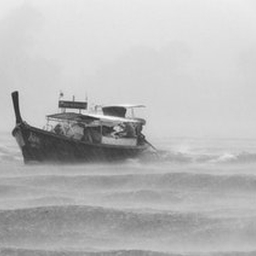}} &
	{\includegraphics[width = 1.17in]{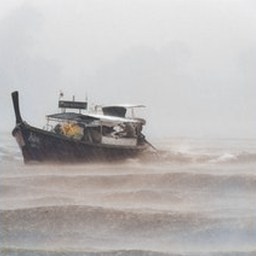}} &
	{\includegraphics[width = 1.17in]{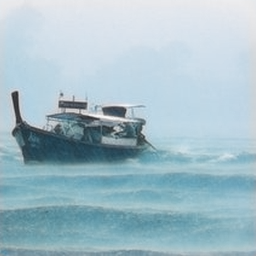}} \\
	{\includegraphics[width = 1.17in]{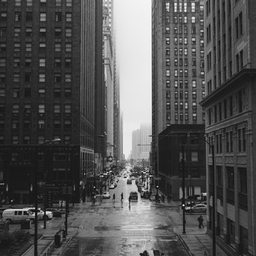}} &
	{\includegraphics[width = 1.17in]{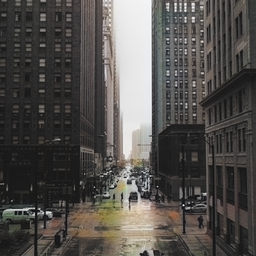}} &
	{\includegraphics[width = 1.17in]{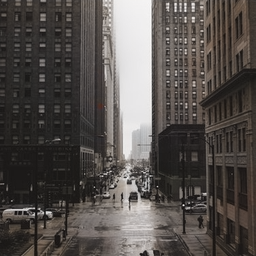}}\\
	{\includegraphics[width = 1.17in]{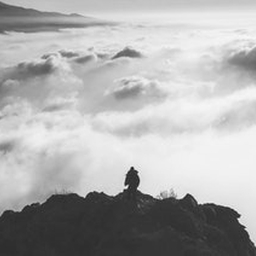}} &
	{\includegraphics[width = 1.17in]{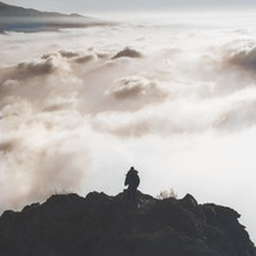}} &
	{\includegraphics[width = 1.17in]{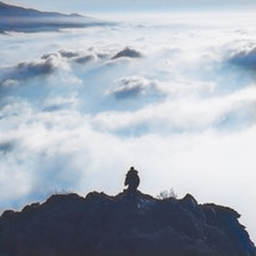}}\\
	{\includegraphics[width = 1.17in]{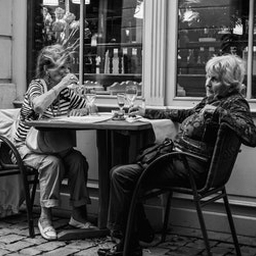}} &
	{\includegraphics[width = 1.17in]{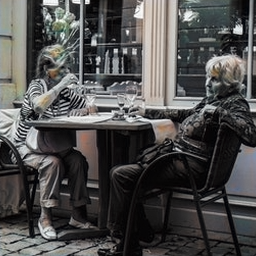}} &
	{\includegraphics[width = 1.17in]{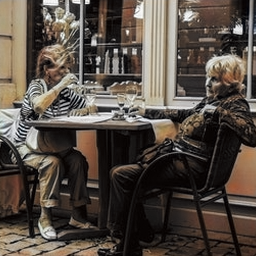}} \\
	{\includegraphics[width = 1.17in]{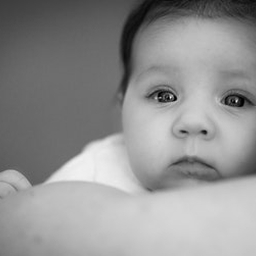}} &
	{\includegraphics[width = 1.17in]{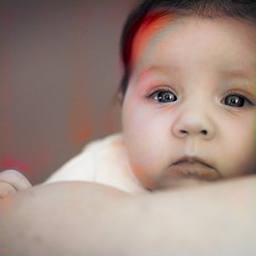}} &
	{\includegraphics[width = 1.17in]{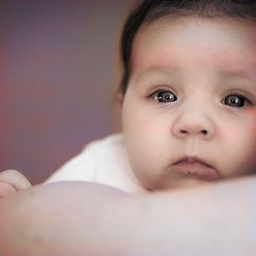}}\\
	Grayscale & ViT-GAN & ViT-I-GAN\\
\end{tabular}
\centering
\caption{Comparing performance of ViT-GAN and ViT-I-GAN}
\label{fig:result}
\end{center}
\end{figure}
As it is observed \ref{fig:result}, results produced by ViT-GAN are not able to colour sky and water which are prominent in those images. It attempts to colourise faces but smears red/orange colour in the process. It faintly colours those images having multiple objects.

In the results produced by ViT-I-GAN, it can be clearly observed that this model colours water and sky appropriately. It also colours faces with much better natural tone without smearing any colour. Additionally, it is able to distinguish between multiple objects and colour them properly.

\subsection{Quantitative  Evaluation}
The metric used for evaluation is Fr\'{e}chet Inception Distance (FID) metric which is used to evaluate quality of the images produced by the generator in a GAN. FID is a measure used to calculate the similarities between images of two datasets, hence compares the distribution of generated images with distribution of real images. Lower FID means better quality of the image and hence a better model. 

It is calculated using the formula:
\\

\centerline{$FID= |\mu -\mu_w|^2  + Tr(\Sigma + \Sigma_w -2(\Sigma\Sigma_w)^{\!1/2})$}

\begin{center}
	
	\begin{tabular}{l}
		$\mu$ and $\mu_w$ represents the feature wise mean of real and generated images.\\

		$\Sigma$ and $\Sigma_w$ are covariance matrix for real and generated feature vectors.\\

		"Tr" is the trace linear algebra operation.\\
\end{tabular}
\end{center}

\begin{table}[!htbp]
\begin{center}
\begin{tabular}{cccc}
\hline
\textbf{} & \textbf{ Epochs } & \textbf{ ViT-GAN } & \textbf{ ViT-I-GAN }\\
\hline
      Unsplash & 25 & 22.73 & 23.06\\
      Unsplash & 50 & 24.89 & 18.16\\
     
\hline
\end{tabular}
\caption{FID comparison on 10k images from Unsplash dataset}
\end{center}
\end{table}

\begin{table}[!htbp]
\centering
\caption{FID comparison on 10k images from COCO dataset}
\begin{center}
\begin{tabular}{cccc}
\hline
\textbf{} & \textbf{ Epochs } & \textbf{ ViT-GAN } & \textbf{ ViT-I-GAN }\\
\hline
      COCO & 50 & 31.2 & 26\\
\hline
\end{tabular}
\end{center}
\end{table}

It was observed that ViT-GAN was performing similar to ViT-I-GAN in terms of FID up till 25 epochs but after that ViT-I-GAN performed much better. This shows that after 25 epochs ViT-GAN stopped improving due to training on limited data whereas the ViT-I-GAN kept on improving demonstrating the usefulness of Inception-v3 embedding in the generator.

\subsection{Comparison on NCD Dataset}
This comparison is based on the Natural-Color Dataset (NCD). These images were chosen for the dataset because they are true to their colour, a banana will most likely be yellowish or greenish whereas colour of sky can range from blue to orangish. The NCD comprises 723 images over 20 different categories. Our models are trained on 400 images and are tested on 323 images. 

We have compared the results of our models with models of various researchers on NCD dataset. In Figure 4, the first column has original coloured images followed by it's corresponding Black \& White image, the next two columns have the output of our models named as ViT-GAN and ViT-I-GAN respectively. The following columns have the output of the similar colourisation work.

The ViT-GAN model is not able to colour the image of strawberry satisfactorily and colours it green. And it also splashes patches of green colour over all images, which is also unwanted. ViT-I-GAN model colourises much better in comparison with the ViT-GAN model. It accurately coloured all fruits with their respective colours and reduced random splashes of colour which were previously observed.

\begin{figure}[!htbp]  
    {\includegraphics[width=1\linewidth]{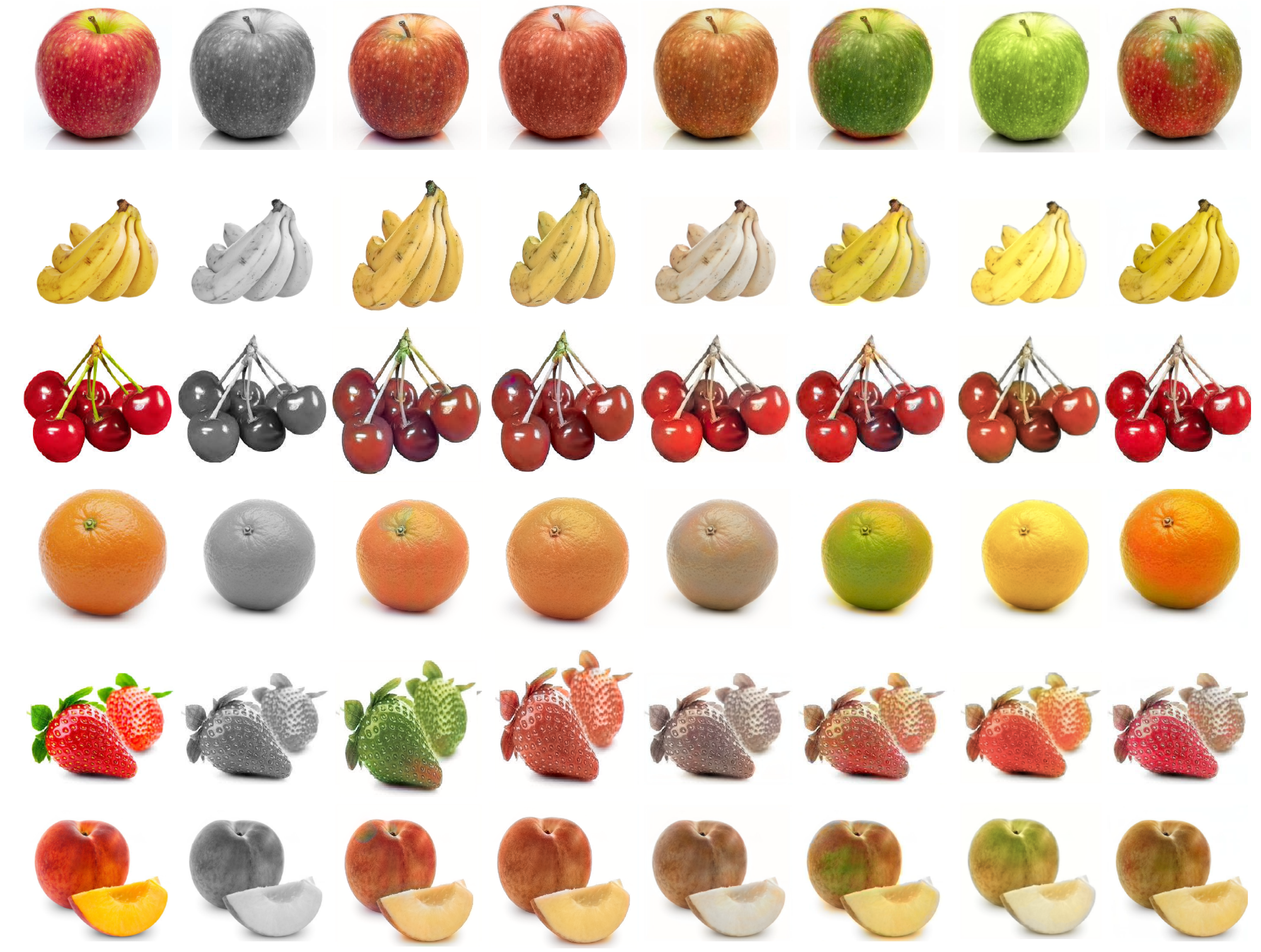} } 
Colored Original \hspace{0.3cm}  B\&W\hspace{0.9cm}  ViT-GAN \hspace{0.5cm} ViT-I-GAN \hspace{0.7cm}      \cite{iiz:sat:edg}  \hspace{1.3cm}   \cite{zha:ric:phi}  \hspace{1.3cm}   \cite{lar:gus:mic}  \hspace{1.3cm}   \cite{zha:ric:jun} \hspace{0.6cm}
	\caption{Comparison on NCD Dataset}
\end{figure}

\begin{center}

	\begin{figure}[!htbp]
	%\hspace*{0.5in}
	\begin{tabular}{cccc}
		{\includegraphics[width=0.2\linewidth]{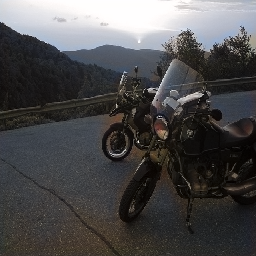}}&
		{\includegraphics[width=0.2\linewidth]{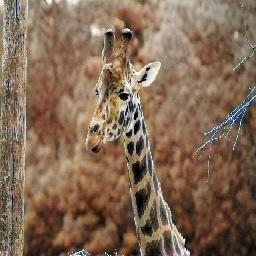}}&
		{\includegraphics[width=0.2\linewidth]{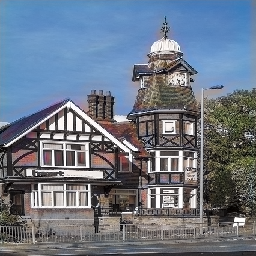}}&
		{\includegraphics[width=0.2\linewidth]{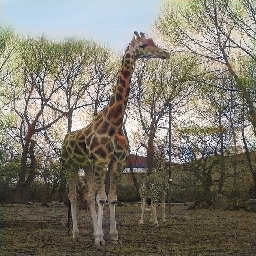}}\\
		{\includegraphics[width=0.2\linewidth]{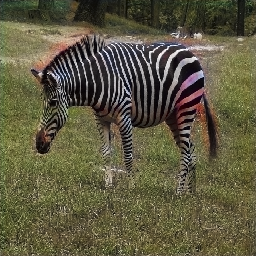}}&
		{\includegraphics[width=0.2\linewidth]{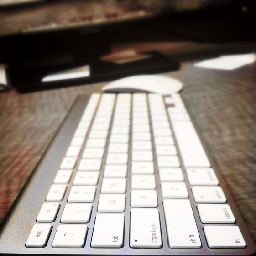}}&
		{\includegraphics[width=0.2\linewidth]{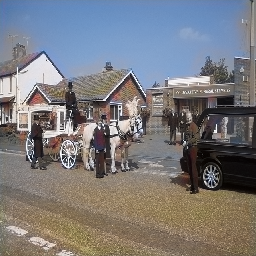}}&
		{\includegraphics[width=0.2\linewidth]{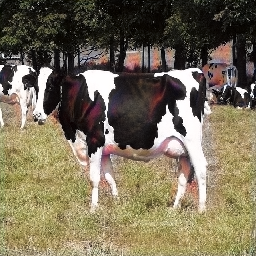}}\\

	\end{tabular}
	\centering
	\caption{Performance of ViT-I-GAN on COCO Test images}
	\label{fig:coco}
	\end{figure}
\end{center}
\subsection{Additional Results}
ViT-I-GAN was trained on the COCO dataset which is a more appropriate dataset for ubiquitous validation. The COCO dataset consists of 118k training images and 40k test images. We have used Adam optimiser with the $\beta_1$ = 0.5 and $\beta_2$ = 0.999. The model was trained for 59k steps with learning rate 2e-4 and then it was trained for 118k steps at a learning rate of 2e-5. 

FID for ViT-I-GAN on COCO test data of 40k images is 23. The model properly colours different objects with their respective colours. It is able to differentiate between zebra, giraffe and cow and colours them appropriately. Hence, training on a dataset like COCO generalises the model which enables it to accurately colour a wide variety of images.

\section{Limitation}
Almost 20 \% of the test images were left uncoloured or barely coloured, due to limited resources on small dataset making the model not able to generalize well on complex images. The model was trained to colourize 256 x 256 images.The models were not validated on the Imagenet dataset.

\section{Conclusion and Future Work}

Our experiments conclude that by fusing InceptionNet-v3 with the generator, the network gets better intuition of various objects in an image. So, the network correlates object representation with colouring schema, especially when the training data is limited. This correlation shows significant improvement over generators not having fusion embedding. This improvement is very noticeable while it colours skies and water and refrains from smearing colours. ViT-I-GAN shows FID improvement of 27\% on train data and 16\% on test data in comparison to ViT-GAN.

Black and White images like historical images, videos and sketches can be colourised better and at scale. 
Further training of ViT-I-GAN on COCO dataset \citep{lin:tsu:mic} resulted in better generalization. This improvement can be further enhanced by using a much larger dataset like ImageNet which contains 1.4M images over 1000 classes which will generalise the model for colouring a wide variety of images. Moreover, by increasing the number of steps for training the model will colourise much better.

%\section{Citations and Bibliographies}
%\bibliographystyle{unsrtnat}
%\bibliography{references}

%\bibliographystyle{unsrtnat}
%\bibliography{references}  %%% Uncomment this line and comment out the ``thebibliography'' section below to use the external .bib file (using bibtex) .

%%% Uncomment this section and comment out the \bibliography{references} line above to use inline references.
% \begin{thebibliography}{1}

% 	\bibitem{kour2014real}
% 	George Kour and Raid Saabne.
% 	\newblock Real-time segmentation of on-line handwritten arabic script.
% 	\newblock In {\em Frontiers in Handwriting Recognition (ICFHR), 2014 14th
% 			International Conference on}, pages 417--422. IEEE, 2014.

% 	\bibitem{kour2014fast}
% 	George Kour and Raid Saabne.
% 	\newblock Fast classification of handwritten on-line arabic characters.
% 	\newblock In {\em Soft Computing and Pattern Recognition (SoCPaR), 2014 6th
% 			International Conference of}, pages 312--318. IEEE, 2014.

% 	\bibitem{hadash2018estimate}
% 	Guy Hadash, Einat Kermany, Boaz Carmeli, Ofer Lavi, George Kour, and Alon
% 	Jacovi.
% 	\newblock Estimate and replace: A novel approach to integrating deep neural
% 	networks with existing applications.
% 	\newblock {\em arXiv preprint arXiv:1804.09028}, 2018.

% \end{thebibliography}

\end{document}